\documentclass[journal, final]{IEEEtran}
\ifCLASSINFOpdf
\else
\fi
\usepackage{graphicx}%
\usepackage{multirow}%
\usepackage{amsmath,amssymb,amsfonts}%

\usepackage{amsthm}%
\usepackage{mathrsfs}%
\usepackage{textcomp}%
\usepackage{manyfoot}%
\usepackage{booktabs}%
\usepackage{algorithm}%
\usepackage{algorithmicx}%
\usepackage{algpseudocode}%
\usepackage{listings}%
\usepackage{array}

\usepackage{float}%
\usepackage{color}%
\usepackage{soul}
\usepackage{tabularx}
\usepackage{tabularray}
\usepackage{hhline}
\usepackage[dvipsnames]{xcolor,colortbl}
\usepackage{makecell}
\usepackage{pifont}
\usepackage{bbding}
\usepackage{arydshln}
\usepackage[pagebackref,breaklinks,colorlinks]{hyperref}
\usepackage{newtxtext}

\begin{document}
%
\title{DFIC: Towards a balanced facial image dataset for automatic ICAO compliance verification}
%
%
%

\author{Nuno~Gonçalves, 
        Diogo~Nunes,
        Carla~Guerra,
        and~João~Marcos
\thanks{N. Gonçalves, D. Nunes and J. Marcos are with the Institute of Systems and Robotics - University of Coimbra, Portugal (e-mail: nunogon@deec.uc.pt; diogo.nunes@isr.uc.pt; joao.marcos@isr.uc.pt).}
\thanks{C. Guerra was with the Institute of Systems and Robotics - University of Coimbra, Portugal (e-mail: carla.guerra@isr.uc.pt).}
}

\maketitle

\begin{abstract}
Ensuring compliance with ISO/IEC and ICAO standards for facial images in machine-readable travel documents (MRTDs) is essential for reliable identity verification, but current manual inspection methods are inefficient in high-demand environments. This paper introduces the DFIC dataset, a novel comprehensive facial image dataset comprising around 58,000 annotated images and 2706 videos of more than 1000 subjects, that cover a broad range of non-compliant conditions, in addition to compliant portraits. Our dataset provides a more balanced demographic distribution than the existing public datasets, with one partition that is nearly uniformly distributed, facilitating the development of automated ICAO compliance verification methods.

Using DFIC, we fine-tuned a novel method that heavily relies on spatial attention mechanisms for the automatic validation of ICAO compliance requirements, and we have compared it with the state-of-the-art aimed at ICAO compliance verification, demonstrating improved results. DFIC dataset is now made public\footnote{\url{https://github.com/visteam-isr-uc/DFIC}} for the training and validation of new models, offering an unprecedented diversity of faces, that will improve both robustness and adaptability to the intrinsically diverse combinations of faces and props that can be presented to the validation system. These results emphasize the potential of DFIC to enhance automated ICAO compliance methods but it can also be used in many other applications that aim to improve the security, privacy, and fairness of facial recognition systems.
\end{abstract}


%
\IEEEpeerreviewmaketitle

\section{Introduction}
\label{sec:intro}
%
%
%
%
\IEEEPARstart{P}{hotographs} used in identity documents, especially in Machine-Readable Travel Documents (MRTDs) such as passports and sovereign identity documents, must meet specific standards to ensure consistency and to enable accurate identification. These standards are enforced through quality metrics that evaluate elements such as a proper head framing, a uniform background, or the use of sunglasses or glasses that cover the eyes, among other criteria.

Three key standardization documents for biometric quality assessment are the ISO/IEC 19794-5 standard \cite{iso19794-5_2011}, the ICAO’s Doc 9303 \cite{doc9303}, and the ICAO Technical Report on Portrait Quality \cite{wolf2018icao}. Together, these recommendations outline criteria for capturing high-quality, passport-style images in MRTDs. 
Examples of such guidelines are the level of contrast of the images (blurred images are not allowed), the absence of objects occluding the face, the obligation to look straight ahead, or uniform lighting. These criteria support the effective operation of Facial Recognition Systems (FRS), which imposes a set of recommendations and restrictions on the photo.
The extensive set of requirements defined by the ISO/IEC standards makes verifying the compliance of a single facial image a challenging task. This process is still largely visual, often relying on human experts with occasional support from automated systems \cite{ferrara2012face}. This reliance on manual review limits efficiency in busy environments, like enrollment offices. Additionally, in some high-demand environments like airports, real-time automatic facial registration of passengers is being used to check in at border gates. In these cases facial images need to be compliant with a subset of 26 requirements. Consequently, there is an ongoing need for full automation of compliance verification, through the development of algorithms that expedite document processing. These algorithms can use conventional image processing techniques, geometric measurements, or advanced methods like deep learning to confirm compliance with the requirements.

\begin{figure}[h]
\centering
    \includegraphics[width=0.8\linewidth]{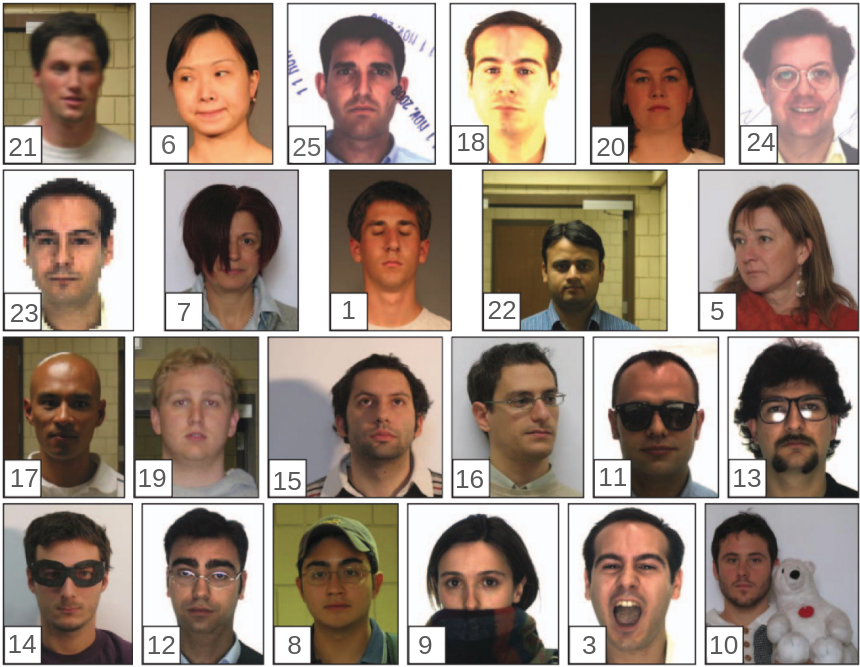}
    \caption{Examples of non-compliant images for the requirements listed in Table \ref{table:req_list} \cite{ferrara2012face}.}
    \label{fig:non_compliant_images}
\end{figure}

Table \ref{table:req_list} compiles the list of 26 photographic and pose-specific requirements according to ISO-IEC 19794-5 standard that are considered in the remainder of this paper. The first 23 requirements were proposed by Biolab \cite{ferrara2012face} and are commonly used in the literature. We opted to add three more, considering the frequency in which they are disregarded (requirements 2, 4 and 26 on the list). Figure \ref{fig:non_compliant_images} shows examples of non-compliant images for the requirements listed in Table \ref{table:req_list}.

The development of methods that accurately verify the ICAO compliance (as this topic is commonly known) of a portrait demands large and varied amounts of data. The publicly available datasets to date do not represent many relevant facial portrait cases that should be validated by ICAO compliance algorithms.

In view of this, we present a novel large facial image dataset entitled Diverse Face Images - Coimbra (DFIC) dataset.
It contains a total of 58,633 images, half taken with a high-quality camera and the other half taken with a lower quality device, and 2706 short videos. The data was collected from more than 1000 subjects and accommodates a large number of the possible situations that make an image non-compliant regarding the ICAO requirements. Each image is annotated according to various attributes regarding the compliance of all the requirements, and when it applies, the reason for the non-compliance and the level of non-compliance. This new dataset provides a distribution across gender, age and origin significantly more balanced than the datasets found in the literature \cite{phillips2005overview}, one order of magnitude above BioLab-ICAO's \cite{ferrara2012face}, which is currently the biggest in this context. The sub-groups considered within age and origin follow the ones considered by the National Institute of Standards and Technology (NIST), particularly for the Face Recognition Technology Evaluation (FRTE) benchmark to Demographic studies \cite{nist_frte},

Additionally, and in order to evaluate how the proposed dataset can enhance the performance of ICAO verification methods, we used the DFIC dataset to train a new method to automatically verify the compliance of facial portraits with the ICAO requirements, achieving state of the art results in all testing scenarios. The results obtained by our solution, when compared with the existing methods, suggest that the DFIC dataset can improve both the accuracy, fairness and generalization ability of ICAO evaluation systems when the models are trained/tuned in the proposed dataset, due to its richness in scale, demographic representation and overall variability. Thus, the DFIC dataset constitutes this work main contribution, by providing a more fair and challenging ICAO compliance verification dataset.



\begin{table}
\small
\centering
\caption{List of facial image quality tests regarding photographic and pose-specific details, according to ISO/IEC 19794-5 standard.}
\label{table:req_list}
\resizebox{=\columnwidth}{!}{%
\begin{tabular}{|l l | l l|}
 \hline
 \textbf{Req.} & \textbf{Test Description} & \textbf{Req.} & \textbf{Test Description} \\
 \hline
 \hline
    1 & Eyes Closed & 14 & Frames Too Heavy\\
    2 & Non-Neutral Expression & 15 & Shadows Behind Head\\
    3 & Mouth Open & 16 & Shadows Across Face \\
    4 & Rotated Shoulders & 17 & Flash Reflection on Skin\\
    5 & Roll/Pitch/Yaw & 18 & Unnatural Skin Tone\\
    6 & Looking Away & 19 & Red Eyes \\
    7 & Hair Across Eyes & 20 & Too Dark/Light\\
    8 & Head Coverings & 21 & Blurred\\
    9 & Veil Over Face & 22 & Varied Background\\
    10 & Other Faces or Toys/Objects & 23 & Pixelation\\
    11 & Dark Tinted Lenses & 24 & Washed Out \\
    12 & Frame Covering the Eyes & 25 & Ink Marked/Creased\\
    13 & Flash Reflection on Lenses & 26 & Posterization \\
 \hline
\end{tabular}
}
\end{table}

\section{Related work}
\label{sec:related_work}

\subsection{Publicly available datasets}

Current publicly available datasets relevant in the field of ICAO compliance lack sufficient diversity in terms of age and ethnicity, as illustrated in Table \ref{tab:table_distributions}. The considered datasets has been chosen based on the commonly usage by the main contributors to ICAO compliance check - Biolab in particular, for their work widely recognized as a standard tool for evaluating ISO/IEC 19794-5 compliance, but also the creators of ICAONet. 
Although constructed long ago, they are the best tailored datasets that are publicly available.

Many datasets do not specify their age and origin distributions. For smaller datasets such as AR \cite{ARdatabase} and PUT \cite{PUT}, we conducted manual annotations. Although this approach may introduce some labeling noise due to the difficulty of accurately inferring age and origin based solely on appearance, it still provides a good estimation of the general demographic overview.
Due to the large number of images in the CelebAMask-HQ dataset \cite{CelebAMaskHQ}, we did not attempt to fully annotate it. Some demographic information, including age and gender, is already available in the literature, though age is categorized simply as 'Young' or 'Not Young.' By examining these annotations, we estimate that the threshold between these labels likely falls within the 30 to 40-years old range. The FRGC dataset \cite{FRGC} includes some demographic information but with limited age diversity. Overall, there is a lack of balance across demographic groups in these datasets.

To address this gap, Biolab recently created a fully synthetic dataset for the evaluation of ICAO compliance methods considering privacy issues. Built on top of the ONOT dataset \cite{onotDataset}, a ICAO-compliant synthetic dataset, the TONO dataset \cite{tonoDataset} addresses subject, photographic and acquisition requirements. Although more balanced across groups, it still lacks scale and representation of numerous requirements considered by the ISO/IEC 19794-5 standard.

\setlength\dashlinedash{0.2pt}
\setlength\dashlinegap{1.5pt}
\setlength\arrayrulewidth{0.3pt}

\begin{table*}[t]
 \caption{Overview regarding gender, origin and age distributions over the existing datasets used in research related to the verification of ICAO requirements. Some of the entries were not specified by the authors - in smaller datasets such as AR and PUT, we manually annotated them; in the high dimension CelebAMask-HQ dataset we marked the entries not annotated with '-'.}
 \label{tab:table_distributions}
	\renewcommand{\arraystretch}{1.0}
	\setlength{\tabcolsep}{5pt}
	\resizebox{\textwidth}{!}{%
		\begin{tabular}{lcc|cc|ccc|c}

            \cmidrule(){1-9} 
             \multirow{2}{*}{\textbf{Dataset}} & \multirow{2}{*}{\textbf{\#Images}} & \multicolumn{1}{c}{\multirow{2}{*}{\textbf{\#Subjects}}} & \multicolumn{2}{c}{\textbf{Gender}} & \multicolumn{3}{c}{\textbf{Origin}} & \multirow{2}{*}{\textbf{Age}} \\

             
              &  &  & \textbf{Male} & \textbf{Female} & \textbf{Asian} & \textbf{White/Caucasian} & \textbf{Other} & \\

             \cmidrule(r){1-1} 
             \cmidrule(lr){2-2} 
             \cmidrule(lr){3-3} 
             \cmidrule(lr){4-5} 
             \cmidrule(lr){6-8} 
             \cmidrule(l){9-9} 

             \textbf{PUT} & 9.971 & 100 & 89\% & 11\% & 0\% & 100\% & 0\% & {[}0-18{]}: 0\%; {[}18-40{]}: 97\%; {[}40+{]}: 3\% \\

             \hdashline

            \textbf{AR} & 3.315 & 136 & 56\% & 44\% & 3\% & 97\% & 0\% & {[}0-18{]}: 1\%; {[}18-40{]}: 74\%; {[}40+{]}: 25\% \\

            \hdashline

            \textbf{TONO+ONOT} & 4.306 & 275 & 61\% & 39\% & 18\% & 49\% & 33\% & {[}18-35{]}: 31\%; {[}36-65{]}: 44\%; {[}66+{]}: 25\% \\

            \hdashline

            \textbf{CelebAMask-HQ} & 30.000 & - & 37\% & 63\% & - & - & - & {[}Young{]}: 78\%; {[}Not Young{]}: 22\% \\

            \hdashline

            \textbf{FRGC} & 40.000 & 466 & 57\% & 43\% & 22\% & 68\% & 10\% & {[}18-22{]}: 65\%; {[}23-27{]}: 18\%; {[}28+{]}: 17\% \\

              \hline

    	\end{tabular}}

\end{table*}

\subsection{Existing methods for automatic ICAO compliance verification}

The University of Bologna's BioLab group has significantly contributed to the advance of methodologies for verifying compliance with the ISO/IEC 19794-5 standard. In 2009, they introduced the BioLab-ICAO framework \cite{maltoni2009biolab}, a pioneering benchmark tool for evaluating face image compliance with ICAO standards. This framework addressed the issues related to the lack of images that fail certain requirements (e.g., the presence of head coverings or sunglasses). In 2012, BioLab expanded the benchmark, integrating images from public datasets - AR \cite{ARdatabase}, FRGC \cite{FRGC}, PUT \cite{PUT}) and proprietary samples. This update resulted in a dataset of 5,588 images from 601 identities, divided into an official benchmark database (4,868 images) and a testing protocol (720 images). Recently, BioLab has further contributed by enhancing available datasets with high-quality synthetic face images \cite{domenico2024}.

BioLab also developed the BioLabSDK \cite{ferrara2012face}, the first framework to define a protocol for evaluating 23 scene requirements for facial image compliance. The proposed approach mostly relies on traditional computer vision techniques with human-engineered features, which present performance limitations in today’s security-focused environments, including lack of generalization. By contrast, recent advancements in Deep Learning have demonstrated superior predictive accuracy and generalization, resulting in less biased and more inclusive compliance verification approaches, despite its relative slower computation.
Currently, the BioLab-ICAO framework is widely recognized as a standard tool for evaluating ISO/IEC 19794-5 compliance through an online public competition, Face Image ISO Compliance Verification (FICV), hosted on the FVC-onGoing platform \cite{fvcongoing}. FICV serves as the primary evaluation tool in research and also for commercial applications, where 23 scene requirements are assessed individually (1-23 in Table \ref{table:req_list}). The results are reported in terms of Equal Error Rate (EER), which is an appropriate and common metric to evaluate the performance of biometric systems that can be defined as the point where the False Acceptance Rate (FAR) and False Rejection Rate (FRR) curves intercept each other.
To date, there are five published algorithms in the FVC-onGoing platform: BioTest \cite{BioTest}, BioPass Face \cite{BioPass}, id3 \cite{id3}, ICAO SDK \cite{icao_sdk} and ICAONet \cite{e2022collaborative}. Their results are directly extracted from the benchmark's published results and are summarized in Table \ref{tab:results_fvc}. In comparison with the BioLabSDK, such algorithms achieve comparable or even better performance rates in some requirements. However, four out of the five algorithms are proprietary solutions owned by private companies, therefore, there is no detailed explanation about their methods. The remainder one - ICAONet \cite{e2022collaborative} - is a deep learning-based method that simultaneously evaluated all 23 requirements, achieving state-of-the-art results in nine of them. It adopts a Multitask Learning (MTL) framework that utilizes a single neural network architecture composed of three core components: an encoder, a decoder, and a dense branch. The encoder and decoder form an auto-encoder; the encoder processes images to generate a feature embedding that supports both unsupervised image reconstruction via the decoder and multi-label classification through the dense branch. This setup provides a comprehensive and efficient solution for compliance verification across all requirements. However, ICAONet’s effectiveness hinges on the quality and diversity of the training data; an adequate dataset can prevent the network from overfitting to specific patterns and make it able to better generalize.

\begin{table}[ht]
\small
\centering
\caption{Results directly extracted from the benchmark's results on the FVC-onGoing platform. Results are reported in terms of EER (in \%) and belong to the five different algorithms submitted.}
\label{tab:results_fvc}
\renewcommand{\arraystretch}{1.1}
\resizebox{=0.75\columnwidth}{!}{%
\begin{tabular}{lccccc}
\hline
\textbf{Req.} &
  \textbf{\begin{tabular}[c]{@{}c@{}}ICAO\\ Net\end{tabular}} &
  \textbf{id3} &
  \textbf{\begin{tabular}[c]{@{}c@{}}BioPass\\ Face\end{tabular}} &
  \textbf{\begin{tabular}[c]{@{}c@{}}Bio\\ Test\end{tabular}} &
  \textbf{\begin{tabular}[c]{@{}c@{}}ICAO\\ SDK\end{tabular}} \\ \hline
1  & 2.1  & 1.7  & \textbf{1.6}  & 30.5 & 48.8 \\ \hdashline
2  & \textbf{5.4}  & 15.3 & 13.3 & 24.2 & 47.5 \\ \hdashline
3  & 49   & -    & 4.8  & \textbf{3.6}  & -    \\ \hdashline
4  & \textbf{1.7}  & 2.1  & 1.9  & 5.1  & 50.0 \\ \hdashline
5  & \textbf{1.2}  & 2.9  & 3.1  & 4.6  & 27.7 \\ \hdashline
6  & 7.3  & 0.2  & \textbf{0}    & 9.2  & -    \\ \hdashline
7  & 29.0 & \textbf{0.2}  & 1.3  & 32.4 & -    \\ \hdashline
8  & 13.7 & -    & 13.0 & \textbf{12.4} & -    \\ \hdashline
9  & 0.8  & \textbf{0.2}  & 4.6  & 6.7  & -    \\ \hdashline
10 & 8.4  & -    & 5.2  & \textbf{3.7}  & 18.7 \\ \hdashline
11 & \textbf{4.6}  & 9.1  & 10.7 & 12.6 & -    \\ \hdashline
12 & \textbf{1.0}  & 1.7  & 1.4  & 1.2  & -    \\ \hdashline
13 & 8.2  & \textbf{1.0}  & 1.7  & 10.3 & 48.3 \\ \hdashline
14 & 3.3  & -    & 5.4  & \textbf{2.4}  & 30.0 \\ \hdashline
15 & \textbf{3.3}  & 10.5 & 9.9  & 15.9 & -    \\ \hdashline
16 & \textbf{0.4}  & 1.0  & 1.8  & 2.1  & 50.0 \\ \hdashline
17 & \textbf{0.8}  & -    & 2.7  & 2.3  & -    \\ \hdashline
18 & 9.5  & \textbf{1.4}  & 2.1  & 3.3  & -    \\ \hdashline
19 & \textbf{2.3}  & 6.6  & 10.7 & 4.0  & -    \\ \hdashline
20 & \textbf{5.7}  & 6.8  & 9.8  & 16.5 & -    \\ \hdashline
21 & \textbf{0}    & -    & 1.4  & 3.7  & 50.0 \\ \hdashline
22 & 2.3  & \textbf{0.6}  & 3.8  & 5.0  & -    \\ \hdashline
23 & 41.4 & -    & \textbf{1.2}  & 15.4 & -    \\ \hline
\end{tabular}
}
\end{table}

Another important recent tool for ICAO compliance verification is the "Open Source Face Image Quality" (OFIQ), an open-source software for assessing the quality of facial images used in face recognition \cite{ofiq}. This tool serves as the reference implementation of algorithms detailed in the revision of ISO/IEC 29794-5 \cite{iso29794-5_tr_2010}, recently published - ISO/IEC 29794-5:2025\cite{iso29794-5:2025}. Designed primarily for use in biometric applications, especially in border control, OFIQ provides a standardized approach to facial image quality assessment. This tool uses simple classical methods to evaluate many requirements (that rely on color spaces, geometric measures or pixel ratios) in a trade-off between accuracy and velocity. As an open-source suite, OFIQ is expected to be continuously developed and improved by the community.

More recently, and resorting to a combination of deep learning and classical methods, BioGaze \cite{bioGazeMethod} introduces an automated framework for ISO/ICAO compliance verification, achieving state-of-the-art results on the synthetic TONO+ONOT dataset. An additional recent work is FaceQVec \cite{hernandez2022faceqvec}, a tool developed by Hernandez-Ortega et al. for assessing facial image compliance with the same 23 requirements defined by BioLab, along with two additional compliance criteria related to white noise estimation and facial expressions. FaceQvec integrates both traditional techniques and pre-trained Convolutional Neural Networks (CNNs) for these evaluations. However, this tool did not undergo BioLab’s benchmark evaluation, nor results were presented for many of the considered requirements, as the authors lacked non-compliant samples both on the development and evaluation databases used.


Despite more than a decade of research on this topic, there are still many open challenges. For instance, some requirements still yield some high EERs as the best result across published studies or providing open-source widely accessible solutions.
A significant limitation is the lack of publicly available datasets specifically tailored to address ICAO compliance. While DL models such as ICAONet have proven valuable to improve the results for ICAO requirements validation, these methods rely on large and varied publicly available datasets to reach their full potential. DFIC dataset is aimed at fulfilling this gap.

\section{DFIC Dataset}
\label{sec:DFIC_dataset}

To overcome the lack of publicly available datasets to develop algorithms to validate the compliance with ICAO requirements we built a large-scale facial image dataset called DFIC.
The dataset contains a total of 58,633 images, taken from 1016 identities gathered across volunteers of different ages, genders and origins. The images were collected in controlled conditions, using many different accessories (around 20) and variations of pose, light (5 types were considered: natural, intense artificial lighting, with and without shadows, smooth artificial lighting, with and without shadows) and expressions (mostly smiling) that did not comply with the ICAO requirements. To the best of our knowledge, this is the largest dataset present in the literature containing images regarding the complete list of ICAO requirements. Half of the images were taken with high quality cameras and the other half were taken using lower quality devices such as smartphones.
It features also 2,706 short videos with a duration of around 5", where the subject performs body, head or face movements (e.g. opening and closing the mouth; rotating the head, among others), which adds several more frames that can be used and that include diverse degrees of variation.

The DFIC dataset is carefully annotated in terms of compliance to ICAO requirements, comprising more than 900,000 manual annotations and 400,000 automatic annotations for the artificial generated images. Additionally, the DFIC dataset also provides around 670,000 manual attribute annotations that describe physical aspects of each image, such as hair characteristics or physical conditions. In total, the dataset provides more than 2,000,000 individual annotations, making it the largest, more diverse and annotated ICAO evaluation dataset to date.  



\subsection{Participants}

\setlength\dashlinedash{0.2pt}
\setlength\dashlinegap{1.5pt}
\setlength\arrayrulewidth{0.3pt}

\begin{table*}[t]
 \caption{Overview of DFIC dataset, regarding gender, origin and age distributions. HQ, SQ and Gen refer to High Quality, Standard Quality and Generated images, respectively.}
 \label{tab:table_distributions2}
	\renewcommand{\arraystretch}{1.1}
	\setlength{\tabcolsep}{7pt}
	\resizebox{\textwidth}{!}{%
		\begin{tabular}{llcc|cc|ccc|ccccc}

            \cmidrule(){1-14} 
             \multicolumn{2}{c}{\multirow{2}{*}{\textbf{Partition}}} & \multirow{2}{*}{\textbf{\#Images}} & \multicolumn{1}{c}{\multirow{2}{*}{\textbf{\#Subjects}}} & \multicolumn{2}{c}{\textbf{Gender}} & \multicolumn{3}{c}{\textbf{Origin}} & \multicolumn{5}{c}{\textbf{Age}} \\

             & &  &  & \textbf{Male} & \textbf{Female} & \textbf{Asian} & \textbf{White/Caucasian} & \textbf{African} & \textbf{{[}0-20{]}} & \textbf{{[}21-35{]}} & \textbf{{[}36-50{]}} & \textbf{{[}51-65{]}} & \textbf{{[}66+{]}} \\

             \cmidrule(r){1-2} 
             \cmidrule(lr){3-3} 
             \cmidrule(lr){4-4} 
             \cmidrule(lr){5-6} 
             \cmidrule(lr){7-9} 
             \cmidrule(l){10-14}

             \multirow{3}{*}{\textbf{DFIC-All}} & HQ & 22.817 & \multirow{3}{*}{1016} & \multirow{3}{*}{52\%} & \multirow{3}{*}{48\%} & \multirow{3}{*}{16\%} & \multirow{3}{*}{68\%} & \multirow{3}{*}{16\%} & \multirow{3}{*}{31\%} & \multirow{3}{*}{36\%} & \multirow{3}{*}{14\%} & \multirow{3}{*}{10\%} & \multirow{3}{*}{9\%} \\

               & SQ & 19.172 & & & & & & & \\
                & Gen & 16.644 & & & & & & & \\

            \hdashline

              \multirow{3}{*}{\textbf{DFIC-Train}} & HQ & 19.996 & \multirow{3}{*}{891} & \multirow{3}{*}{52\%} & \multirow{3}{*}{48\%} & \multirow{3}{*}{14\%} & \multirow{3}{*}{72\%} & \multirow{3}{*}{14\%} & \multirow{3}{*}{33\%} & \multirow{3}{*}{38\%} & \multirow{3}{*}{13\%} & \multirow{3}{*}{8\%} & \multirow{3}{*}{8\%} \\

               & SQ & 17.179 & & & & & & & \\
                & Gen & 14.624 & & & & & & & \\

            \hdashline

              \multirow{3}{*}{\textbf{\begin{tabular}[l]{@{}l@{}}DFIC-Train \\ Balanced\end{tabular}}} & HQ & 8.788 & \multirow{3}{*}{388} & \multirow{3}{*}{49\%} & \multirow{3}{*}{51\%} & \multirow{3}{*}{32\%} & \multirow{3}{*}{36\%} & \multirow{3}{*}{32\%} & \multirow{3}{*}{21\%} & \multirow{3}{*}{22\%} & \multirow{3}{*}{20\%} & \multirow{3}{*}{20\%} & \multirow{3}{*}{17\%} \\

               & SQ & 6.704 & & & & & & & \\
                & Gen & 6.185 & & & & & & & \\

            \hdashline

              \multirow{3}{*}{\textbf{DFIC-Test}} & HQ & 2821 & \multirow{3}{*}{125} & \multirow{3}{*}{50\%} & \multirow{3}{*}{50\%} & \multirow{3}{*}{34\%} & \multirow{3}{*}{35\%} & \multirow{3}{*}{31\%} & \multirow{3}{*}{21\%} & \multirow{3}{*}{22\%} & \multirow{3}{*}{19\%} & \multirow{3}{*}{19\%} & \multirow{3}{*}{19\%} \\

               & SQ & 1.993 & & & & & & & \\
                & Gen & 2.020 & & & & & & & \\

              \hline

    	\end{tabular}}

\end{table*}

To address the lack of diversity in current publicly available datasets (see Table \ref{tab:table_distributions}), in DFIC dataset we collected a diverse set of facial images, ensuring representation across a wide range of ages, from toddlers to seniors, as well as various ethnic backgrounds (see Table \ref{tab:table_distributions2}). We classified participants by age, gender, and origin, paying particular attention to groups typically underrepresented in the available datasets, such as children and seniors. To fill this noticeable gap, we made additional efforts to include more individuals from these demographic groups.
%
Regarding the gender attribute we considered only two classes - Male (M) or Female (F).
The age attribute includes 5 groups - Children/Teen (age between 0 and 20 years), Young Adult (age between 21 and 35 years), Adult (age between 36 and 50 years), Senior Adult (age between 51 and 65 years) and Senior (age 66 years or above). These groups were defined following NIST standards, particularly for the FRTE benchmark to Demographic studies \cite{nist_frte}.
The origin of the participants was divided into 3 groups - 'Caucasian', 'African' and 'Asian' - although one can fine-grain it if needed because, during the dataset acquisition, the country of origin of each identity was registered. Some datasets used in this context use the term 'White' to refer to the group we called 'Caucasians' - we tried to choose the term that sounded more inclusive and respectful to us, although cultural differences may require different terms. The same reasoning applies to other terms used.

We emphasize that the main differences between DFIC Dataset and the most complete current publicly available one - FRGC, include the fact that DFIC has not only more subjects (more than twice), but also it is more balanced across all demographics groups, with subjects of varied age groups, and an extensive list of annotations that may help new algorithms and models to improve their performance.

\subsection{Images collected}

\begin{figure}[h!]
\centering
  \includegraphics[width=0.75\linewidth]{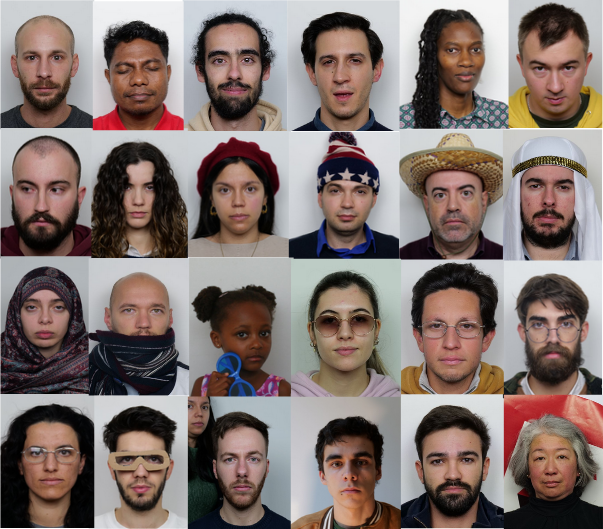}
  \caption{24 sample pictures taken on the dataset.}
  \label{fig:dataset}
\end{figure}

Images were collected in single sessions with each participant, drawing from a varied pool that included students, researchers, professors, staff, and their families at University of Coimbra, as well as a local secondary school. Additional participants came from diverse communities living in our city (mainly African), local African festivals, a shopping mall, a day center for the elderly, and international conferences. Further sampling was conducted in specific neighborhoods and in the Center-East of India, where one of our research team members was temporarily based.

Each volunteer was asked to take pictures in 24 scenarios based on ICAO's standards:
\begin{itemize}
    \item 2 fully compliant scenarios with the ICAO requirements;
    \item 2 potentially compliant scenarios, depending on the MRTD issuer policy on head coverings and glasses;
    \item 20 non compliant with at least one of requirements.
\end{itemize}

All the scenarios included taking two pictures - one with a high quality camera and another with a lower quality device such as a smartphone. Additionally, a short video from each participant was also collected, with an approximate duration of 5" where the subject performs body, head or face movements (e.g. opening and closing the mouth; rotating the head up, down, to left, to the right, among others), which adds many more frames that can be used and that include diverse degrees of variation.
In the end, the DFIC dataset is composed of:
\begin{itemize}
    \item 58,633 facial images;
    \item 2,706 short videos of 5".
\end{itemize}

An example of the complete list of pictures taken by each participant can be seen in Figure \ref{fig:dataset}. Besides the 23 requirements considered by BioLab, we consider another 3 that we find crucial considering the frequency in which they are disregarded. The latter are: having a 'Non-neutral Expression', having 'Rotated Shoulders' and 'Posterization' effects.
It is also worth noting that 16,644 images were artificially generated by applying unnatural skin tone, red eyes, pixelation, washed-out, ink-marked/creased and posterization effects to mostly compliant images from DFIC dataset in order to be able to address requirements \#18, \#19, \#23, \#24, \#25 and \#26. Additionally, we also recurred to artificial image generation to increase the variability for the following requirements: 'Too Dark/Light' through under and over exposition manipulation; 'Blurred' through gaussian blur at several degrees; and 'Varied Background' through the substitution of the background. These generated images are only used for the model training and testing of the respective requirements.

The proportion of compliant and non-compliant samples is summarized per requirement in Table \ref{tab:compliance_dist}. Note that some requirements cannot be verified in some images (e.g. open eyes when the subject wears dark tinted glasses). Consequently, the rows of Table \ref{tab:compliance_dist} not always sum up equally. This was a particularly relevant aspect that we had in mind when building the DFIC dataset in order to make sure that enough samples on both classes were always present.

The image capture setup varied slightly depending on the many places where the acquisition sessions took place.
Most of the pictures were collected in indoor rooms with controlled artificial lighting, including two softboxes and a small portable light to create shadows on purpose. A big white panel right behind each subject was also used, mainly to guarantee an homogeneous background when needed. Only a small portion of the pictures were taken with uncontrolled lighting conditions.
The subjects were positioned between 1.5 to 2 meters from the camera that was kept in a tripod.
To collect the high quality images and the short videos, two different cameras were used - a Canon EOS 350D with a 18-55mm zoom lens and a Panasonic Lumix S5 with a 20-60mm zoom lens.
The lower quality images were taken using lower quality devices such as smartphones (an iPhone 6 and a Samsung Galaxy S7).


\begin{table}
\caption{Compliance (C) \textit{versus} non-compliance (NC) distribution of the DFIC dataset images regarding each ICAO requirement. $\dagger$ To overcome the lack on the number or variability of NC images for some of the requirements, we generated a total of 16644 artificial non-compliant images.}
\label{tab:compliance_dist}
\renewcommand{\arraystretch}{1.1}
\centering
\resizebox{=1.0\columnwidth}{!}{
\begin{tabular}{lcccc}
\hline
\textbf{Requirement} &                \textbf{Rq\#} & \textbf{C} & \textbf{NC} & \textbf{\% NC} \\ \hline
Eyes Closed   & 1              & 37915      & 2522        & 6,24\%         \\ \hdashline
Non-Neutral Expression  & 2    & 38214      & 2232        & 5,55\%         \\ \hdashline
Mouth Open     & 3             & 37742      & 3740        & 9,02\%         \\ \hdashline
Rotated Shoulders   & 4        & 39871      & 1715         & 4,12\%         \\ \hdashline
Roll/Pitch/Yaw    & 5          & 37142      & 4703        & 11,24\%         \\ \hdashline
Looking Away    & 6            & 35920      & 1995        & 5,26\%         \\ \hdashline
Hair Across Eyes   & 7         & 40756      & 1201         & 2,86\%         \\ \hdashline
Head Coverings     & 8         & 32938      & 9042        & 21,54\%        \\ \hdashline
Veil Over Face    & 9          & 40119      & 1864         & 4,44\%         \\ \hdashline
Other Faces or Toys/Objects & 10 & 38209     & 3774        & 8,99\%         \\ \hdashline
Dark Tinted Lenses  & 11        & 40104      & 1873         & 4,46\%         \\ \hdashline
Frame Coverings the Eyes  & 12  & 39479      & 2462        & 5,87\%         \\ \hdashline
Flash Reflection on Lenses  & 13 &39262      & 2717        & 6,47\%         \\ \hdashline
Frames Too Heavy   & 14         & 39922      & 2057         & 4,90\%         \\ \hdashline
Shadows Behind Head  & 15       & 39315      & 2668        & 6,35\%         \\ \hdashline
Shadows Across Face   & 16      & 38702      & 3159        & 7,55\%         \\ \hdashline
Flash Reflection on Skin & 17   & 40768      & 1188         & 2,83\%         \\ \hdashline
Unnatural Skin Tone  & 18       & 41915      & 3 + 3937 $\dagger$        & 8,59\%         \\ \hdashline
Red Eyes  & 19                  & 37620      & 799 $\dagger$          & 2,08\%         \\ \hdashline
Too Dark/Light  & 20            & 39907      & 1632  + 1992 $\dagger$      & 8,33\%         \\ \hdashline
Blurred     & 21                & 37677      & 4300 + 996 $\dagger$        & 12,32\%         \\ \hdashline
Varied Background   & 22        & 36796      & 5183 + 2944 $\dagger$       & 18,09\%         \\ \hdashline
Pixelation    & 23              & 41983      & 996 $\dagger$        & 2,32\%        \\ \hdashline
Washed Out    & 24              & 41983      & 996 $\dagger$       & 2,32\%        \\ \hdashline
Ink Marked/Creased   & 25       & 41983      & 2988 $\dagger$       & 6,64\%        \\ \hdashline
Posterization    & 26           & 41983      & 996 $\dagger$       & 2,32\%        \\ \hline
\end{tabular}
}
\end{table}

As for the videos, they are of interest for future methods, discussed in the Sec. \ref{sec:conclusion}. Due to space limitations we don't include experiments explicitly relying on them.

\subsection{Annotation strategy}

The high-quality images in the DFIC dataset were manually annotated by two independent annotators who were specially trained for this task. An annotation tool was provided to facilitate the process, where annotators were responsible for assessing each image's compliance with all the 26 ICAO requirements considered. The available options were:
\begin{itemize}
    \item 'Compliant': The ICAO requirement is met;
    \item 'No Way to Confirm': Compliance could not be verified due to factors such as poor lighting or occlusion, or there was a high degree of uncertainty about the image compliance;
    \item 'Non-compliant': Specific options were provided to describe the reason for non-compliance (e.g. in the case of shadows behind head, the annotators could distinguish between strong or soft shadows; or, in the case of head coverings, which type of head covering is present).
\end{itemize}

In the cases where the labels conflicted, a third expert annotator evaluated the image and determined the final label. Such strategy led to a very low labeling error dataset.

Additional annotations with respect to specific face and hair details were also conducted. They include the presence of eye or skin conditions, makeup, false eyelashes, earrings, facial piercings, facial hair and the hair size, type and color. 
Each image label also includes information regarding the gender, age and origin of the corresponding subject. However, once the data is made available, the research community can conduct more detailed analyses. In Supplementary Material we present the extensive list of annotations.

\subsection{Legal aspects}

We believe that granting authorized access to the DFIC dataset for research purposes is essential to boost the development of improved solutions for automatic verification of ICAO requirements, among other applications within the scope of improving the security, privacy and fairness of FRS. To facilitate this access, we have adhered to all legal requirements. Each individual featured in the photographs has provided explicit written consent for the use of their image rights for the purpose of improving FRS, and has been informed about our privacy policy that outlines how their images will be processed.
To protect individuals' rights, each photo is tagged with a unique ID assigned to the corresponding person, implementing a pseudonymization technique that restricts access to their names and contact details for dataset users. Our established protocol limits access to biographical data to a select few individuals, who can only use it for the purpose of deleting or editing details at the request of the individuals involved. Furthermore, we require all dataset users to comply with applicable data protection laws, including the general data protection regulation (GDPR) \cite{gdpr2016}. All related documents have been thoroughly reviewed by a team of legal professionals, and we obtained a positive legal assessment from both Institute of Systems and Robotics - Coimbra's DPO and Ethical Committee of the University of Coimbra, under the statement CEIUC$\_$11/R-9$\_$20241211. 

\begin{figure*}[tb]
	\centering\
	\includegraphics[width=0.9\textwidth]{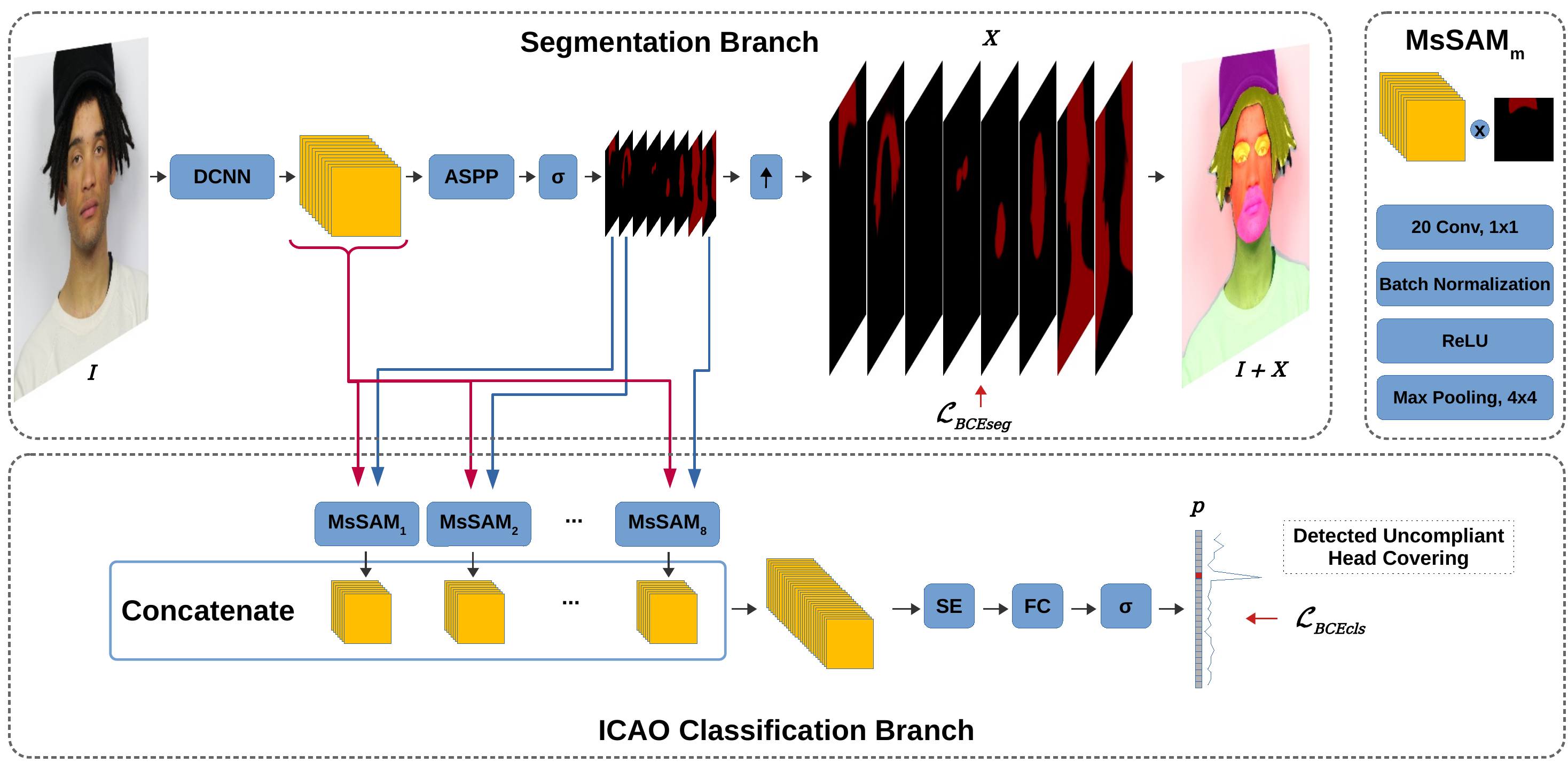}
    \caption{General architecture of the proposed method. It can be decomposed in two main branches: a segmentation branch that recurs to the DeepLabV3 architecture, resorting to a MobileNetV3 Deep Convolutional Neural Network (DCNN) as feature encoder and Atrous Spatial Pyramid Pooling (ASPP) for semantic segmentation; and a classification branch, that resorts to Mask specific Spatial Attention Modules (MsSAMs), a channel attention module through Squeeze and Excitation (SE) and a Fully Connected (FC) layer for final classification. Finally, the $\sigma$ and $\uparrow$ blocks correspond to a sigmoid activation function and an up-sampling operation, respectively.} 
	\label{fig:diagram}
\end{figure*}

\section{Method}
\label{sec:icao_algorithms}

This section details the methodology employed to assess the compliance of face images according to the ISO/IEC 19794-5 standard. We have followed a single model strategy for all requirements, which allows for a fast and complete ICAO compliance evaluation with a single inference. Given the high number of requirements, the developed model presents a strong focus on attention mechanisms in order to guide the learning procedure to extract meaningful features and prevent overfitting. As can be seen in the general architecture in Figure \ref{fig:diagram}, we make use of the DeepLabV3 \cite{deeplabv3} architecture to extract encoded features and facial segmentation maps to be used as spatial attention mechanisms. In summary, the entire framework can be decomposed in two branches: a face region segmentation branch, that recurs to the DeepLabV3 architecture; and a ICAO prediction branch, which applies spatial and channel attention mechanisms for ICAO requirements classification. Resorting to the Mask specific Spatial Attention Mechanisms (MsSAMs) modules, where a point-wise multiplication of the extracted features with each of the segmentation maps occurs, the DCNN is guided to extract relevant and spatial dependent features for each part of the face image. After all MsSAM computations, the region specific features are concatenated and passed through a Squeeze and Excitation (SE) \cite{squeezeE} module to act as a channel attention mechanism, where the most relevant channels, and therefore regions, are enhanced with the resulting encoded features. For the final prediction, a Fully Connected (FC) layer is used to compute the final compliance scores.

For the segmentation/attention maps, eight regions were chosen ($M=8$): (1) head coverings; (2) hair; (3) eyeglasses; (4) eyes; (5) mouth; (6) full face; (7) torso; and (8) background. Since the presented dataset does not provide segmentation maps for model supervision, we extracted most of the mask labels, $Y$, resorting to the FaRL face parser \cite{farl}, whilst the eye and mouth region masks were extracted using the SPIGA landmark detection framework \cite{spiga} to have an estimated region even if these regions are occluded by sunglasses, hair or clothing. Unlike most face parsers, where each pixel belongs to only one class, we employ a multi label prediction mechanism resorting to Binary Cross Entropy (BCE) Loss, and thus, allowing for overlapping segmentation masks. Finally, and given that some face regions are substantially smaller than others, e.g., eyes region when compared to the torso region in a $H \times W$ image, a positive weight factor, $\lambda_{m}$, is also added to the BCE Loss in order to ensure equal importance among different region masks, and prevent negligence of smaller region masks:

\begin{equation}
\begin{aligned}
	\mathcal{L}_{BCEseg} = \frac{-1}{MHW} \sum_{m}^{M} (\lambda_{m} Y_m \cdot log(X_m)+ \\
    + (1-Y_m)\cdot log(1 - X_m))\label{eq:1}
\end{aligned}
\end{equation} 

Similarly, also the ICAO prediction branch is supervised in a multilabel classification fashion, allowing for multiple non-compliances to be triggered in a single image. We also recur to a positive weight factor to account for the natural imbalance that exists between compliant and non-compliant ICAO requirement labels. This weight factor, $\lambda_{r}$ is calculated in order to provide equal importance between compliant and non-compliant labels in the same requirement. Additionally, a $\beta_{r}$ is also used in order to balance the representation across $R$ different requirements:

\begin{equation}
\begin{aligned}
	\mathcal{L}_{BCEcls} = \frac{-1}{R} \sum_{r}^{R} \beta_{r} \cdot g(r,t)\cdot(\lambda_{r} t_r \cdot log(p_r)+ \\
    + (1-t_r)\cdot log(1 - p_r))\label{eq:2}
\end{aligned}
\end{equation} 
where $g(r,t)$ represents a gate function that returns a binary response, $0$ or $1$, depending on the requirement being analyzed, $r$, and the image compliance labels, $t$. The main purpose of this function is to suppress the backward gradient of certain requirements when there are conflicting non-compliances in a given image, e.g., to not consider the loss of closed/opened eyes when the subject is using dark tinted lenses.

Finally, the landmarks retrieved by SPIGA \cite{spiga} can be used to further refine the detection of non-frontal gaze if a more fine decision is required. Based on the distance between the detected iris/pupil and the eye center location for both eyes, a simple, yet effective gaze metric is obtained. This extra computation is only calculated if both frontal gaze and frontal face pose requirements are classified as compliant by the proposed single model, and only operates for the detection of small deviations of the gaze yaw relatively to the center.

\section{Results}
\label{sec:results}

In this section, we present the experimental results conducted on different testing scenarios. Particularly, the results on the proposed DFIC dataset; the results on the TONO+ONOT dataset in order to test the generalization capabilities of the solutions and their respective training domains; and lastly, the results on demographic group bias of each solution resorting to the DFIC dataset.

\subsection{ICAO compliance evaluation on DFIC dataset}

In order to evaluate the relevance of the proposed dataset and method, we first present the results of the proposed method (\textbf{Ours}) on the test partition of the DFIC dataset, and compare them with other publicly available solutions, specifically the ICAONet \cite{e2022collaborative}, the BioGaze \cite{bioGazeMethod} and the OFIQ \cite{ofiq}.

\begin{table}[bpt]
\caption{Performance comparison in terms of EER on the test partition of the DFIC dataset.}
\label{table:results_facing}
\renewcommand{\arraystretch}{1.3}
\setlength{\tabcolsep}{3pt}
\centering
\resizebox{=1.0\columnwidth}{!}{
\begin{tabular}{lccccc}
\hline
\textbf{Requirement} &   \textbf{ICAONet} &    \textbf{ICAONet*} & \textbf{BioGaze} & \textbf{OFIQ} & \textbf{Ours} \\ \hline
Eyes Closed   & 0.094   &  0.028    & 0.065      & 0.009        & \textbf{0.008}         \\ \hdashline
Non-Neutral Expression  & -  & 0.130  & 0.224      & 0.064        & \textbf{0.025}         \\ \hdashline
Mouth Open     & 0.315    & 0.123   & -      & 0.019        & \textbf{0.016}         \\ \hdashline
Rotated Shoulders   & -   & 0.108    & 0.202      & -         & \textbf{0.020}         \\ \hdashline
Roll/Pitch/Yaw    & 0.210  & 0.108   & 0.229      & 0.042        & \textbf{0.027}         \\ \hdashline
Looking Away    & 0.293    & 0.143     & 0.466      & -        & \textbf{0.139}         \\ \hdashline
Hair Across Eyes   & 0.397  & 0.128   & -      & -         & \textbf{0.026}         \\ \hdashline
Head Coverings     & 0.227   & 0.153   & 0.102      & 0.089        & \textbf{0.020}        \\ \hdashline
Veil Over Face    & 0.349   & 0.030    & -      & -         & \textbf{0.002}         \\ \hdashline
Other Faces or Toys/Objects & 0.520  & 0.114 & -      & -        & \textbf{0.007}         \\ \hdashline
Dark Tinted Lenses  & 0.104   & 0.011     & 0.175      & 0.054         & \textbf{0}         \\ \hdashline
Frame Coverings the Eyes  & 0.241  &  0.029 & -      & -        & \textbf{0}        \\ \hdashline
Flash Reflection on Lenses  & 0.140 & 0.067 & -      & -        & \textbf{0.025}         \\ \hdashline
Frames Too Heavy   & 0.248    & 0.015     & -      & -         & \textbf{0}         \\ \hdashline
Shadows Behind Head  & 0.335    & 0.091   & -      & -        & \textbf{0.033}         \\ \hdashline
Shadows Across Face   & 0.410   & 0.039  & 0.305     & 0.237        & \textbf{0.011}         \\ \hdashline
Flash Reflection on Skin & 0.116  & 0.083  & -      & -         & \textbf{0.017}         \\ \hdashline
Unnatural Skin Tone  & 0.358  & 0.014  & 0.280      & 0.033         & \textbf{0}         \\ \hdashline
Red Eyes  & 0.427          & 0.023   & -      & -          & \textbf{0.005}         \\ \hdashline
Too Dark/Light  & 0.121     &0.064    & 0.144      & 0.129         & \textbf{0.045}         \\ \hdashline
Blurred     & 0.292       & 0.095    & 0.133      & 0.243        & \textbf{0.046}         \\ \hdashline
Varied Background   & 0.317   & 0.063   & 0.120      & 0.227        & \textbf{0.028}         \\ \hdashline
Pixelation    & 0.445     & 0.049    & 0.214      & -        & \textbf{0}        \\ \hdashline
Washed Out    & 0.466      &  0.029    & 0.279      & -       & \textbf{0.003}        \\ \hdashline
Ink Marked/Creased   & 0.492   &  0.114   & -      & -      & \textbf{0.006}        \\ \hdashline
Posterization   & -    &  0.029    & 0.172      & -       & \textbf{0}        \\ \hline 
\textbf{Average}   & 0.301  &  0.072      & 0.207      & 0.104       & \textbf{0.020}        \\ \hline
\end{tabular}
}
\end{table}

As depicted in Table \ref{table:results_facing}, our approach achieves the best performance scores across all the requirements. However, it is an expected behavior, since \textbf{Ours} method was trained using the train partition of the same dataset used for testing, avoiding domain shifts that may occur when applying other methods. To study the superiority of the proposed method, disregarding domain shift influence, a fine-tuning was performed on ICAONet using the proposed dataset and the results were also presented in Table \ref{table:results_facing} (column ICAONet*). ICAONet was chosen for fine-tuning given its similarities with \textbf{Ours} method: both are single model deep learning classification systems, which in turn enables a fine-tuning for different domains and datasets. As can be observed in Table \ref{table:results_facing}, that presents the two versions of ICAONet, we managed to decrease the average EER from 0.301 to 0.072, but still with a value substantially higher than the EER obtained by the proposed method (0.020), and thus, suggesting that the superiority of \textbf{Ours} method is not only due to the domain factor, but also to the efficiency of the method itself.

\subsection{ICAO compliance evaluation on TONO+ONOT dataset}

Another important answer that we aim for is whether in fact, the proposed dataset enables generalization capabilities to the methods tuned for it. To test this hypothesis, we resorted to the TONO+ONOT dataset \cite{tonoDataset}, which consists of a fully synthetic dataset for the development and evaluation of ISO/ICAO face image compliance systems. With a shorter list of requirements, Biolab has only considered a total of 14 requirements. To the best of our knowledge none of the compared methods, with the exception of the BioGaze framework, was tuned or validated to this dataset, making it an ideal candidate to test the generalization capabilities of the studied methods, and consequently also the domains where these methods were trained/tuned.

From the results presented in Table \ref{table:results_tono}, we can conclude that \textbf{Ours} method trained on the proposed dataset achieves the lowest overall EER, and thus, suggesting that the proposed dataset can in fact provide the generalization capability required to train and evaluate ISO/ICAO compliance systems. Additionally, and looking specifically at the results obtained by ICAONet on the TONO dataset, we can see that when trained on the proposed dataset we manage to decrease the overall EER from 0.429 to 0.155, confirming once again the generalization gain provided by the DFIC dataset. Although this decrease is substantial, the EER is still far superior than the one obtained by \textbf{Ours} approach, suggesting that to obtain a reliable and generalized solution, the dataset used for model tuning and validation is not enough, and that the method itself plays a crucial role in the final performance.


\begin{table}
\caption{Performance in terms of EER on the TONO+ONOT dataset. The column DFIC$_{req}$ corresponds to the equivalent requirement or set of requirement indexes considered in the DFIC dataset.} 
\label{table:results_tono}
\renewcommand{\arraystretch}{1.3}
\setlength{\tabcolsep}{3pt}
\centering
\resizebox{=1.0\columnwidth}{!}{
\begin{tabular}{lcccccc}
\hline
\textbf{Requirement} & \textbf{DFIC$_{req}$} &  \textbf{ICAONet} & \textbf{ICAONet*} & \textbf{BioGaze} & \textbf{OFIQ} & \textbf{Ours} \\ \hline

Head w/o coverings  & 8 & 0.175   & 0.113    & 0.003      & \textbf{0}        & \textbf{0}        \\ \hdashline
Gaze in camera & 6 & 0.445 & 0.399   & \textbf{0.032}      & -        & 0.125         \\ \hdashline
Eyes open    & 1 & 0.125    & 0.061     & 0.016      & \textbf{0}        & 0.003         \\ \hdashline
No/light makeup  & - & -   & -  & \textbf{0.002}      & -         & -         \\ \hdashline
Neutral expression & 2 & 0.339  & 0.203    & 0.093      & 0.039        & \textbf{0.035}         \\ \hdashline
No sunglasses  &  11 & 0.027  & 0.045     & 0.014      & \textbf{0}        & \textbf{0}         \\ \hdashline
Frontal pose  & 4, 5 & 0.234   & 0.434   & 0.093      & 0.171        & \textbf{0.042}         \\ \hdashline
Correct exposure   &  20 & 0.358  & 0.183    & 0.105      & 0.335        & \textbf{0.100}        \\ \hdashline
In focus photo  &  21 & 0.368   & 0.010  & \textbf{0}      & \textbf{0}         & \textbf{0}         \\ \hdashline
Correct saturation & 18 & 0.321 & 0.019 & 0.041     &   0.172      & \textbf{0.003}         \\ \hdashline
Uniform background  & 22 & 0.417  & 0.051    & 0.117      & 0.297         & \textbf{0.023}         \\ \hdashline
Uniform face lighting  & 16, 20 & 0.376 & 0.257 & \textbf{0.085}      & 0.110        & 0.132       \\ \hdashline
No pixelation & 23 & 0.494 & 0.215 & \textbf{0}      & -        & 0.006         \\ \hdashline
No posterization  & 26 & -  & 0.023   & 0.008      & -         & \textbf{0}         \\ \hline
\textbf{Average} &   & 0.429   & 0.155    & 0.044      & 0.113       & \textbf{0.036}        \\ \hline
\end{tabular}
}
\end{table}

\subsection{Study on demographic group bias}

A crucial aspect of face image classification systems is their performance bias when applied in different demographic groups. Although the test partition of the proposed dataset is already balanced across demographic groups, which in turn exposes the possible solution's demographic bias, we further inspect the fairness of each method by conducting an isolated evaluation for each group, and compare the performance increase/decrease inside each demographic category, as shown in Table \ref{tab:demo_results}.

\begin{table*}[btp]
 \caption{Performance comparison regarding demographic group bias. The \textbf{Overall} column refers to the EER of all testing data. For each demographic group and respective column, a \textcolor{Green}{-val} refers to a decrease of \textbf{val} in the overall EER for that particular group, meaning a better performance, while a \textcolor{Red}{+val} to an increase and consequently worse performance. The last column, \textbf{Bias Index} refers to a metric that reflects how prone each method is to demographic group bias, and is obtained by computing the maximum pairwise difference in each category (Gender, Origin and Age), and finally performing their sum. A lower value of this index indicates that a given method is more resilient to demographic group bias.}
 \label{tab:demo_results}
	\renewcommand{\arraystretch}{1.1}
	\resizebox{\textwidth}{!}{%
		\begin{tabular}{lc|cc|ccc|ccccc|c}

            \cmidrule(){1-13} 
             \multirow{2}{*}{\textbf{Model}} & \multicolumn{1}{c}{\multirow{2}{*}{\textbf{Overall}}} & \multicolumn{2}{c}{\textbf{Gender}} & \multicolumn{3}{c}{\textbf{Origin}} & \multicolumn{5}{c}{\textbf{Age}} & \multicolumn{1}{c}{\multirow{2}{*}{\textbf{Bias Index}}} \\

             & & \textbf{Male} & \textbf{Female} & \textbf{Asian} & \textbf{Caucasian} & \textbf{African} & \textbf{{[}0-20{]}} & \textbf{{[}21-35{]}} & \textbf{{[}36-50{]}} & \textbf{{[}51-65{]}} & \textbf{{[}66+{]}} & \\

             \cmidrule(r){1-1} 
             \cmidrule(lr){2-2} 
             \cmidrule(lr){3-4} 
             \cmidrule(lr){5-7} 
             \cmidrule(lr){8-12} 
             \cmidrule(l){13-13} 

             \textbf{ICAONet} & 0.301 & \textcolor{Red}{+0.023} & \textcolor{Green}{-0.017} & \textcolor{Green}{-0.002} & \textcolor{Green}{-0.039} & \textcolor{Red}{+0.013} & \textcolor{Green}{-0.006} & \textcolor{Green}{-0.017} & 0.000 & \textcolor{Green}{-0.028} & \textcolor{Red}{+0.054} & 0.174
             \\ \hdashline

             \textbf{ICAONet*} & 0.072 & \textcolor{Red}{+0.001} & \textcolor{Red}{+0.004} & \textcolor{Red}{+0.004} & \textcolor{Green}{-0.029} & \textcolor{Red}{+0.018} & \textcolor{Green}{-0.004} & \textcolor{Green}{-0.024} & \textcolor{Green}{-0.012} & \textcolor{Green}{-0.007} & \textcolor{Red}{+0.034} & 0.108
             \\ \hdashline

             \textbf{ICAONet* - Bal} & 0.079 & \textcolor{Green}{-0.003} & \textcolor{Green}{-0.002} & \textcolor{Red}{+0.012} & \textcolor{Green}{-0.023} & \textcolor{Red}{+0.025} & \textcolor{Green}{-0.011} & \textcolor{Green}{-0.011} & \textcolor{Green}{-0.005} & \textcolor{Green}{-0.020} & \textcolor{Red}{+0.023} & 0.092
             \\ \hdashline

             \textbf{Biogaze} & 0.207 & \textcolor{Green}{-0.019} & \textcolor{Green}{-0.028} & \textcolor{Green}{-0.033} & \textcolor{Green}{-0.005} & \textcolor{Red}{+0.019} & \textcolor{Green}{-0.007} & \textcolor{Green}{-0.001} & \textcolor{Green}{-0.025} & \textcolor{Green}{-0.023} & \textcolor{Red}{+0.004} & 0.090
             \\ \hdashline

             \textbf{OFIQ} & 0.104 & \textcolor{Green}{-0.019} & \textcolor{Red}{+0.005} & \textcolor{Green}{-0.003} & \textcolor{Green}{-0.006} & \textcolor{Red}{+0.013} & \textcolor{Green}{-0.002} & \textcolor{Green}{-0.009} & \textcolor{Green}{-0.019} & \textcolor{Green}{-0.034} & \textcolor{Red}{+0.012} & 0.089
             \\ \hdashline

             \textbf{Ours} & \textbf{0.020} & \textcolor{Red}{+0.002} & \textcolor{Red}{+0.002} & \textcolor{Red}{+0.005} & \textcolor{Green}{-0.013} & \textcolor{Green}{-0.002} & \textcolor{Red}{+0.001} & \textcolor{Green}{-0.012} & \textcolor{Green}{-0.004} & \textcolor{Green}{-0.010} & \textcolor{Red}{+0.017} & 0.047\\ \hdashline

             \textbf{Ours - Bal} & 0.024 & \textcolor{Red}{+0.001} & \textcolor{Green}{-0.001} & \textcolor{Red}{+0.003} & \textcolor{Green}{-0.009} & \textcolor{Green}{-0.004} & \textcolor{Green}{-0.009} & \textcolor{Green}{-0.007} & \textcolor{Green}{-0.006} & \textcolor{Green}{-0.009} & \textcolor{Red}{+0.015} & \textbf{0.038}\\

              \hline

    	\end{tabular}}

\end{table*}

For demographic group bias measurement, we designed an index to reflect how biased a method is towards a given group in deterioration of others in the same demographic category. This metric, named Bias Index, takes the EER difference between the best and worst performing group in each category (Gender, Origin and Age), and performs their sum to obtain the final value. If $EER_{c}^{max}$ represents the EER of the demographic group with the maximum value in a given category $c$ (and $EER_{c}^{min}$ the corresponding minimum), then the bias index can be formulated as:

\begin{equation}
\begin{aligned}
	Bias~Index = \sum_{c} EER_{c}^{max} - EER_{c}^{min}\label{eq:3}
\end{aligned}
\end{equation} 
where $c\in\{Gender,~Origin,~Age\}$. Additionally, and in order to obtain more insights into demographic bias behavior, two additional models were added to this analysis, ICAONet*-Bal and Ours-Bal. These models refer to the methods trained on the balanced train partition of the DFIC dataset, and serve the purpose of inspecting the influence that a balanced training domain in terms of representation has on the final solution fairness.

From the results in Table \ref{tab:demo_results} we can conclude that both versions of \textbf{Ours} method achieve the lowest values of demographic bias index, indicating a higher degree of fairness, while also providing the best overall performance across ICAO requirements. Regarding the remaining models, ICAONet presents the highest demographic bias, but with a substantial decrease when trained on the proposed dataset, achieving a bias index similar to OFIQ and BioGaze. Additionally, when comparing the results in demographic bias between the solutions trained in the balanced and non-balanced partitions of DFIC dataset (comparison between ICAONet* and ICAONet*-Bal; and between Ours and Ours-Bal), one can notice a decrease in the bias index with the cost of overall ICAO performance. This behavior indicates that a balanced training domain in terms of representation can in fact improve the solutions' fairness, however with a slight cost in overall performance due to lesser training data.

Finally, by inspecting the results on particular demographic groups, we can conclude that all methods suffer a performance decrease in the senior ([66+]) age group, which suggests that this bias is not only due to lack of representation, but also because it comprises a harder classification task. The reason behind it may be due to physical aspects of this age group, such as affected posture or lowered eyelids, which can hinder both the models accuracy and the objectivity of the labeling process. Another demographic group that suffers an increased error across methods is the African one. With the exception of Ours method, which strongly relies on attention mechanisms, all models revealed a negative bias towards this group, which may not only be due to lack of representation on training domain but also due to the lower contrast caused by the darker skin tone, affecting proper ICAO requirements evaluation. Hence, we can conclude that in order to achieve low demographic bias in ICAO requirements evaluation systems, a balanced representation across demographics is not sufficient, and that the method itself also plays a fundamental role to obtain fair classification systems.

\section{Conclusion}
\label{sec:conclusion}

Portraits in identity documents, particularly on Machine-Readable Travel Documents (MRTDs) such as passports, must adhere to strict quality standards to facilitate reliable identification of their holders. To address the inefficiency of manual review, not only in the enrollment phase but also at some specific scenarios, as seamless facial recognition, there is a growing need for automated compliance algorithms that ensure faster and more consistent document processing.

This paper presents a new, extensive, facial image dataset, the DFIC dataset, designed to support automated ICAO compliance verification. The dataset can also be used in many other applications intended to improve the security, privacy, and fairness of FRS.
The dataset comprises 58,633 images {\textendash} captured with both high and low-quality cameras  {\textendash} and 2,706 short videos from over 1000 subjects, covering a broad range of non-compliant conditions across different genders, ages, and origins. Each image is annotated for compliance with specific requirements, providing a significantly larger and more balanced resource than the existing available datasets.
Using the DFIC dataset, we fine-tuned a novel method with strong focus on spatial attention mechanisms for the specific task of validating the compliance of the 26 ICAO requirements in a single inference. The developed model achieved SoTA results in all the testing scenarios, while also achieving the best results regarding fairness and demographic bias. Our experiments show that models trained on DFIC achieve enhanced accuracy in realistic scenarios due to dataset’s variety and complexity, setting a new dataset for ICAO compliance verification.

In the future, we plan to submit the developed method to the FvC-Ongoing benchmark conducted by Biolab \cite{fvcongoing} to evaluate its performance on an additional dataset that is particularly relevant in the context of ICAO requirements.
Additionally, we plan to collect more images to enhance the DFIC dataset (focusing in particular on older female subjects from African and Asian origins) to achieve a fully balanced representation. Finally, experiments with the acquired videos are also planned, particularly the development of an application that can output the best video frame in terms of ICAO compliance, which would greatly benefit the process of automatic enrollment of users in some specific contexts. Additionally, videos can also help in improving some of the verification algorithms, since they can provide multiple images with intermediate states, and not only images in the extreme of compliance (e.g. for the requirement Mouth Open, a video can provide the algorithm with several states of the mouth, that may improve its decision capacity).


%



\section*{Acknowledgment}

The authors would like to thank the Portuguese Mint and Official Printing Office (INCM) and the Institute of Systems and Robotics - University of Coimbra for the support of the project FACING2. This work has been supported by Fundacão para a Ciência e a Tecnologia (FCT) under the project UIDB/00048/2020.

\ifCLASSOPTIONcaptionsoff
  \newpage
\fi



%


\bibliographystyle{IEEEtran}
\bibliography{Transactions-Bibliography/IEEEexample}

@String(CVPR= {IEEE Conf. Comput. Vis. Pattern Recog.})

@String(BMVC= {Brit. Mach. Vis. Conf.})

@String(ICIP = {IEEE Int. Conf. Image Process.})

@String(CVPR  = {CVPR})

@String(BMVC  =	{BMVC})

@String(ICIP  = {ICIP})

@article{e2022collaborative,
  title={{A collaborative deep multitask learning network for face image compliance to ISO/IEC 19794-5 standard}},
  author={Arnaldo Silva and Herman Gomes and Leonardo Batista},
  journal={{Expert Systems with Applications}},
  volume={198},
  pages={116756},
  year={2022},
  publisher={Elsevier}
}

@inproceedings{hernandez2022faceqvec,
  title={{FaceQvec: Vector quality assessment for face biometrics based on ISO compliance}},
  author={Javier Hernandez-Ortega and Julian Fierrez and Luis F. Gomez and Aythami Morales and Jose Gonzalez-de-Suso and Francisco Zamora-Martinez},
  booktitle={{Proceedings of the IEEE/CVF Winter Conference on Applications of Computer Vision}},
  pages={84--92},
  year={2022}
}

@misc{nist_frte,
  author = {{National Institute of Standards and Technology}},
  title = {{Face Technology Evaluations - FRTE/FATE}},
  howpublished = {\url{https://www.nist.gov/programs-projects/face-technology-evaluations-frtefate}},
  year = 2024,
  note = {Accessed: 2024-05-24}
}

@misc{BioTest,
  author = {{BioTest}},
  title = {{Result of algorithm BioTest 1.3.8 on FICV-1.0}},
  howpublished = {\url{https://biolab.csr.unibo.it/FvcOnGoing/UI/Form/AlgResult.aspx?algId=2787}},
  year = 2014,
  note = {Accessed: 2024-10-02}
}

@misc{BioPass,
  author = {{BioPassFace}},
  title = {{Result of algorithm BioPass Face 5.6 on FICV-1.0}},
  howpublished = {\url{https://biolab.csr.unibo.it/FvcOnGoing/UI/Form/AlgResult.aspx?algId=6336}},
  year = 2017,
  note = {Accessed: 2024-10-02}
}

@misc{id3,
  author = {{ICAO\_compliance}},
  title = {{Result of algorithm ICAO\_compliance 1.1.4 on FICV-1.0}},
  howpublished = {\url{https://biolab.csr.unibo.it/FvcOnGoing/UI/Form/AlgResult.aspx?algId=5343}},
  year = 2016,
  note = {Accessed: 2024-10-02}
}

@misc{icao_sdk,
  author = {{ICAO SDK}},
  title = {{Result of algorithm ICAO SDK 1.0.0 on FICV-1.0}},
  howpublished = {\url{https://biolab.csr.unibo.it/FvcOnGoing/UI/Form/AlgResult.aspx?algId=8134}},
  year = 2021,
  note = {Accessed: 2024-10-02}
}

@misc{fvcongoing,
  author = {{BioLab}},
  title = {{FVC-ongoing BioLab Benchmark}},
  howpublished = {\url{https://biolab.csr.unibo.it/fvcongoing/UI/Form BenchmarkAreas/BenchmarkAreaFICV.aspx9}},
  year = 2000,
  note = {Accessed: 2023-06-30}
}

@inproceedings{iso29794-5_tr_2010,
  author = {{ISO/IEC}},
  title={{ISO/IEC TR 29794-5:2010
Information technology — Biometric sample quality — Part 5: Face image data}},
  booktitle={},
  journal={{ISO standards}},
  year={2010}
}

@inproceedings{iso29794-5:2025,
  author = {{ISO/IEC}},
  title={{ISO/IEC 29794-5:2025
Information technology — Biometric sample quality — Part 5: Face image data}},
  booktitle={},
  journal={{ISO standards}},
  year={2025}
}

@inproceedings{iso19794-5_2011,
  author = {{ISO/IEC 19794-5:2011}},
  title={{ISO/IEC 19794-5:2011
Information technology — Biometric data interchange formats — Part 5: Face image data}},
  booktitle={},
  journal={{ISO standards}},
  year={2011}
}

@inproceedings{doc9303,
  author = {ICAO}, 
  title={{Doc 9303 - Machine Readable Travel Documents - Part 3: Specifications Common to all MRTDs - Eighth Edition}},
  journal={{ICAO Doc Series}},
  year={2021}
}

@inproceedings{maltoni2009biolab,
  title={{Biolab-icao: A new benchmark to evaluate applications assessing face image compliance to iso/iec 19794-5 standard}},
  author={Davide Maltoni and Annalisa Franco and Matteo Ferrara and Dario Maio and Antonio Nardelli},
  booktitle={{2009 16th IEEE International Conference on Image Processing (ICIP)}},
  pages={41--44},
  year={2009},
  organization={IEEE}
}

@article{ferrara2012face,
  title={{Face image conformance to iso/icao standards in machine readable travel documents}},
  author={Matteo Ferrara and Annalisa Franco and Dario Maio and Davide Maltoni},
  journal={{IEEE Transactions on Information Forensics and Security}},
  volume={7},
  number={4},
  pages={1204--1213},
  year={2012},
  publisher={IEEE}
}

@article{wolf2018icao,
  author={A. Wolf},
  title={{ICAO: Portrait Quality (Reference Facial Images for MRTD), Version 1.0. Standard}},
  journal={{International Civil Aviation Organization}},
  year={2018}
}

@article{ARdatabase,
author = {A. Martinez and Robert Benavente},
year = {1998},
month = {01},
title = {{The AR face database}},
journal = {{Tech. Rep. 24 CVC Technical Report}}
}

@INPROCEEDINGS{FRGC,
  author={P.J. Phillips and P.J. Flynn and T. Scruggs and K.W. Bowyer and Jin Chang and K. Hoffman and J. Marques and Jaesik Min and W. Worek},
  booktitle={{2005 IEEE Computer Society Conference on Computer Vision and Pattern Recognition (CVPR'05)}}, 
  title={{Overview of the face recognition grand challenge}}, 
  year={2005},
  volume={1},
  number={},
  pages={947-954 vol. 1}
}

@INPROCEEDINGS{PUT,
author = {Kasiński, Andrzej and Florek, A and Schmidt, Adam},
year = {2008},
month = {01},
pages = {59-64},
title = {The PUT face database},
volume = {13},
journal = {Image Processing and Communications}
}

@inproceedings{CelebAMaskHQ,
  title={Maskgan: Towards diverse and interactive facial image manipulation},
  author={Lee, Cheng-Han and Liu, Ziwei and Wu, Lingyun and Luo, Ping},
  booktitle={Proceedings of the IEEE/CVF conference on computer vision and pattern recognition},
  pages={5549--5558},
  year={2020}
}

@inproceedings{domenico2024,
  title={{ONOT: a High-Quality ICAO-compliant Synthetic Mugshot Dataset}},
  author={Nicolò Domenico and Guido Borgui and Annalisa Franco and Davide Maltoni},
  booktitle={{Proceedings of the 18th IEEE International Conference on Automatic Face and Gesture Recognition}},
  year={2024}
}

@misc{ofiq,
  author = {{Federal Office for Information Security}},
  title = {{Open Source Face Image Quality
(OFIQ) – Draft Report}},
  howpublished = {\url{https://www.bsi.bund.de/SharedDocs/Downloads/EN/BSI/OFIQ/Projektabschlussbericht_OFIQ_1_0.pdf?__blob=publicationFile&v=6}},
  year = 2022,
  note = {Accessed: 2024-08-27}
}

@inproceedings{phillips2005overview,
  title={Overview of the face recognition grand challenge},
  author={Phillips, P Jonathon and Flynn, Patrick J and Scruggs, Todd and Bowyer, Kevin W and Chang, Jin and Hoffman, Kevin and Marques, Joe and Min, Jaesik and Worek, William},
  booktitle={2005 IEEE computer society conference on computer vision and pattern recognition (CVPR'05)},
  volume={1},
  pages={947--954},
  year={2005},
  organization={IEEE}
}

@misc{gdpr2016,
  title        = "{Regulation (EU) 2016/679 of the European Parliament and of the Council of 27 April 2016 on the protection of natural persons with regard to the processing of personal data and on the free movement of such data (General Data Protection Regulation)}",
  howpublished = "\textit{Official Journal of the European Union}, L119, 1--88",
  year         = 2016,
  month        = may,
  note         = "Available at: \url{https://eur-lex.europa.eu/eli/reg/2016/679/oj}",
}

@article{deeplabv3,
  title={Rethinking Atrous Convolution for Semantic Image Segmentation},
  author={Liang-Chieh Chen and George Papandreou and Florian Schroff and Hartwig Adam},
  journal={ArXiv},
  year={2017},
  volume={abs/1706.05587},
  url={https://api.semanticscholar.org/CorpusID:22655199}
}

@inproceedings{farl,
  title={General facial representation learning in a visual-linguistic manner},
  author={Zheng, Yinglin and Yang, Hao and Zhang, Ting and Bao, Jianmin and Chen, Dongdong and Huang, Yangyu and Yuan, Lu and Chen, Dong and Zeng, Ming and Wen, Fang},
  booktitle={Proceedings of the IEEE/CVF Conference on Computer Vision and Pattern Recognition},
  pages={18697--18709},
  year={2022}
}

@inproceedings{spiga,
  author    = {Andrés  Prados-Torreblanca and José M Buenaposada and Luis Baumela},
  title     = {Shape Preserving Facial Landmarks with Graph Attention Networks},
  booktitle = {33rd British Machine Vision Conference 2022, {BMVC} 2022, London, UK, November 21-24, 2022},
  publisher = {{BMVA} Press},
  year      = {2022},
  url       = {https://bmvc2022.mpi-inf.mpg.de/0155.pdf}
}

@INPROCEEDINGS{squeezeE,
  author={Hu, Jie and Shen, Li and Sun, Gang},
  booktitle={2018 IEEE/CVF Conference on Computer Vision and Pattern Recognition}, 
  title={Squeeze-and-Excitation Networks}, 
  year={2018},
  volume={},
  number={},
  pages={7132-7141},
  doi={10.1109/CVPR.2018.00745}
}

@INPROCEEDINGS{bioGazeMethod,
  author={Elatfi, Osama and Domenico, Nicolò Di and Borghi, Guido and Franco, Annalisa and Maltoni, Davide},
  booktitle={2025 IEEE 19th International Conference on Automatic Face and Gesture Recognition (FG)}, 
  title={BioGaze: a Framework for Evaluating the Photographic Requirements of the ISO/IEC 39794-5 Standard}, 
  year={2025},
  volume={},
  number={},
  pages={1-10},
  doi={10.1109/FG61629.2025.11099423}
}

@inproceedings{tonoDataset,
  title={TONO: a Synthetic Dataset for Face Image Compliance to ISO/ICAO Standard},
  author={Borghi, Guido and Franco, Annalisa and Di Domenico, Nicol{\`o} and Maltoni, Davide},
  booktitle={The 18th European Conference on Computer Vision Workshops 2024},
  year={2024}
}

@inproceedings{onotDataset,
title={ONOT: a High-Quality ICAO-compliant Synthetic Mugshot Dataset},
author={Di Domenico, Nicol{`o} and Borghi, Guido and Franco, Annalisa and Maltoni, Davide and others},
booktitle={The 18th IEEE International Conference on Automatic Face and Gesture Recognition (FG)},
pages={1--6},
year={2024}
}

\end{document}